\documentclass[conference]{IEEEtran}
\usepackage{times}

\usepackage[numbers]{natbib}
\usepackage{multicol}
\usepackage[bookmarks=true]{hyperref}

\usepackage[ruled,vlined,linesnumbered]{algorithm2e}
\usepackage{amsmath,amssymb,amsfonts}
\usepackage{subcaption}
\usepackage{algorithmic}
\usepackage{graphicx}
\usepackage{color}

\newcommand{\m}{^{\mbox{\scriptsize{-}}}}

\newtheorem{theorem}{Theorem}
\newtheorem{lemma}[theorem]{Lemma}
\newtheorem{proposition}[theorem]{Proposition}

\newenvironment{proof}{\textit{Proof.}}{{\leavevmode\nobreak\hfil\penalty50\hskip0em\vadjust{}
		\nobreak\hfil$\Box$\parfillskip=0pt\finalhyphendemerits=0\par}\vspace{1ex} }

\begin{document}

\title{Belief Space Planning: A Covariance\\ Steering Approach}

% You will get a Paper-ID when submitting a pdf file to the conference system

% avoiding spaces at the end of the author lines is not a problem with
% conference papers because we don't use \thanks or \IEEEmembership

\author{\authorblockN{Dongliang Zheng\authorrefmark{1},
Jack Ridderhof\authorrefmark{1},
Panagiotis Tsiotras\authorrefmark{1}, and
Ali-akbar Agha-mohammadi\authorrefmark{2}}
\authorblockA{\authorrefmark{1}
Georgia Institute of Technology,
Atlanta, Georgia 30332--0150, USA\\ \{dzheng, jridderhof3, tsiotras\}@gatech.edu}

\authorblockA{\authorrefmark{2}NASA-JPL, California Institute of Technology,
Pasadena, CA 91109 USA\\ aliagha@jpl.nasa.gov}}

\maketitle

\begin{abstract}
A new belief space planning algorithm, called covariance steering Belief RoadMap (CS-BRM), is introduced, which is a multi-query algorithm for motion planning of dynamical systems under simultaneous motion and observation uncertainties.
CS-BRM extends the probabilistic roadmap (PRM) approach to belief spaces and is based on the recently developed theory of covariance steering (CS) that enables guaranteed satisfaction of terminal belief constraints in finite-time.
The nodes in the CS-BRM are sampled in belief space and represent distributions of the system states. 
A covariance steering controller steers the system from one BRM node to another, thus acting as an edge controller of the corresponding belief graph that ensures belief constraint satisfaction.
After the edge controller is computed, a specific edge cost is assigned to that edge. 
The CS-BRM algorithm allows the sampling of non-stationary belief nodes, and thus is able to explore the velocity space and find efficient motion plans.
The performance of CS-BRM is evaluated and compared to a previous belief space planning method, demonstrating the benefits of the proposed approach.
\end{abstract}

\IEEEpeerreviewmaketitle

\section{Introduction}

Motion uncertainty and measurement noise arise in all real-world robotic applications.
When evaluating the safety of a robot under motion and estimation uncertainties, it is no longer sufficient to rely only on deterministic indicators of performance, such as whether the robot is in collision-free or in-collision status. 
Instead, the state of the robot is best characterized by a probability distribution function (pdf) over all possible states, which is commonly referred to as the \textit{belief} or
information state~\cite{Bonet2000Planning,Thrun2005Probabilistic,Van2012Motion}.
Explicitly taking into account the motion and observation uncertainties thus requires planning in the belief space, which allows one to compute the collision probability and thus make more informed decisions.
Planning under motion and observation uncertainties is referred to as belief space planning, 
which can be formulated as a partially observable Markov decision process (POMDP) problem \cite{Kaelbling1998Planning}.
Solving POMDP in general domains, however, is very challenging, despite some recent progress in terms of more efficient POMDP solvers \cite{Ong2010Planning, ross2008online, shani2013survey, somani2013despot, sun2020stochastic}. 
% In planning problems with incomplete information of the state, one needs to keep track of the whole distribution of all possible system  states, referred to as the belief or information state.
Planning in infinite-dimensional distributional (e.g., belief) 
spaces can become more tractable by the use of roadmaps, that is, graphs constructed by sampling.
Since their introduction~\cite{alterovitz2007stochastic, Prentice2009The}, such
belief roadmaps (BRMs) have increased in popularity owing to their simplicity and their ability to avoid local minima.

Sampling-based motion planning algorithms such as probabilistic roadmaps (PRM) \cite{Kavraki1996Prob} and rapidly exploring random trees (RRTs) \cite{Karaman2011Sampling} 
can be used to solve planning problems 
in high-dimensional continuous state spaces, by building a roadmap or a tree incrementally 
through sampling the search space.
%The PRM* and RRT* algorithms introduced in \cite{Karaman2011Sampling} are asymptotically optimal
%versions of the PRM and RRT algorithms, respectively.
%Depending on the available time or computational power, a plan is found, which balances time/computational power with optimality.
However, traditional PRM-based methods only address deterministic systems. 
%deterministic systems are considered in these works \cite{Kavraki1996Prob, Karaman2011Sampling, marble2013asymptotically, gammell2014informed}. 
PRM methods have been extended to belief space planning using  belief space roadmaps (BRMs)~\cite{Prentice2009The, Agha-Mohammadi2014FIRM, Van2011LQG-MP}.
One of the main challenges of belief space roadmap (BRM) methods is that the costs of different edges depend on each other, resulting in the ``curse of history'' problem for POMDPs \cite{Pineau2003Point, Agha-Mohammadi2014FIRM}.
This dependence between edges breaks the optimal substructure property, which is required for search algorithms such as Dijkstra's algorithm or A*.
This problem arises from the unreachability of the belief nodes; even if the robot has full control of its mean, it is difficult to reach 
higher order moments (e.g., 
a specified covariance).
Since the nodes in BRM are sampled in the belief space, the edges in BRM should ideally steer the robot from one distribution to another.
If reachability of BRM nodes is not achieved, an edge in the BRM depends on all preceding edges along the path.

% The \textit{belief} may be classified into estimation belief, control belief, and full belief, which will be denoted by \textit{e-belief}, \textit{c-belief}, and \textit{f-belief}, respectively. 
% \textit{e-belief} is the belief of the state estimation error, \textit{c-belief} is the belief of the estimated state, and \textit{f-belief} is the belief of the actual state.
% \textit{e-belief} planning tries to obtain better state estimation and finds a path with minimum estimation uncertainty.
% \textit{f-belief} is the combination of \textit{e-belief} and \textit{c-belief}.
% If the robot has full knowledge of its state, then \textit{c-belief} is equivalent to \textit{f-belief}. 
% In this paper, we consider the \textit{f-belief} planning problem and will call it \textit{belief} planning for simplicity.  

In general, in the literature, the \textit{belief} may be classified into \textit{(i)} the estimation belief (e-belief), describing the output of the estimator, i.e., the pdf over the error between the estimated value and true value of the state; \textit{(ii)} control belief (c-belief), which  refers to the pdf over ``separated control" error, i.e., the error between the estimated state and the desired state; and \textit{(iii)} the full belief (f-belief), that is, the belief over the true state. 
We discuss these different beliefs in greater detail in Section~\ref{Sec:CovSteering}. 
We just note here that \textit{e-belief} planning tries to obtain better state estimates and finds a path with minimum estimation uncertainty. 
The \textit{f-belief} planning aims at minimizing the full uncertainty that takes into account the impact of the controller as well. 
In this paper, we consider the \textit{f-belief} planning problem and will call it \textit{belief} planning for simplicity.

Planning in \textit{e-belief} space was studied in~\cite{Prentice2009The, Bopardikar2016Robust}, where the goal was to find the minimum estimation uncertainty path for a robot from a starting position to a goal position. 
In \cite{Van2011LQG-MP}, a linear-quadratic Gaussian (LQG) controller was used for motion planning within the BRM context.
By taking into account the controller and the sensors used, \cite{Van2011LQG-MP} computes the true \textit{a priori} probability distribution of the state of the robot. 
Thus, \cite{Van2011LQG-MP} studies the \textit{f-belief} planning problem.
However, the independence between edges is not satisfied.
Similarly, \cite{Bry2011rapidly} studies the \textit{f-belief} planning problem using a tree.
% Planning under intermittent sensing is studied in \cite{Bopardikar2016Robust}. The cost function is chosen as a bound on the maximum eigenvalue of estimation error covariance of a Kalman filter. However, the control problem is not considered. Similar to \cite{Prentice2009The}, the estimation error covariance is only a measure of the confidence of the state estimation, and is insufficient to characterize the uncertainty of state.

The state-of-the-art in terms of BRM methods is probably \cite{Agha-Mohammadi2014FIRM}, which tackles the ``curse of history'' problem. 
The proposed SLQG-FIRM method achieves node reachability using a stationary LQG controller. 
One limitation of this method is that the nodes have to be stationary.
That is, the nodes in the BRM graph need to be sampled in the equilibrium space of the robot, which usually means zero velocity.
Thus, this method cannot explore the velocity space and the resulting paths are suboptimal.
Secondly, a converging process is required at every node.
That is, the robot will have to ``wait'' at each node, which will increase the time required for the robot to reach the goal.
Some remedies are introduced in \cite{Agha-Mohammadi2012Periodic} for systems (e.g., a fixed-wing aircraft) that cannot reach zero velocity by using periodic trajectories and periodic controllers. 
The periodic controller is applied repeatedly until the trajectory of the vehicle converges to the periodic trajectory.
Thus, a ``waiting'' procedure is still required in these approaches.
Online replanning in belief space is studied in \cite{agha2018slap}. The method in \cite{agha2018slap} 
improves the online phase by recomputing the local plans, which includes adding a virtual belief node and local belief edges to the FIRM graph, and solving a dynamic program at every time-step. The offline roadmap construction phase is the same as FIRM \cite{Agha-Mohammadi2014FIRM}. 

% In terms of offline BRM construction, the FIRM method is directly used without modification. Thus, we refer to the FIRM as the state-of-the-art of the BRM method. In this paper, we focus on belief space roadmap algorithms and aim to improve the FIRM method itself. The rollout-based replanning idea presented in \cite{agha2018slap} may be combined with our CS-BRM method to further improve the online planning performance.

Recent developments in explicitly controlling the covariance of a linear system~\cite{Chen2015Optimal, Chen2015Optimal2} 
provide an appealing approach to construct the BRM with guarantees of node reachability.
In particular, for a discrete-time linear stochastic system, covariance steering theory designs a controller that steers the system from an initial Gaussian distribution to a terminal Gaussian distribution in finite-time \cite{Bakolas2016Optimal, Goldshtein2017Finite, Okamoto2018Optimal}. 
Reference \cite{Chen2015Optimal} formulated the finite-horizon covariance control problem as a stochastic optimal control problem. 
In \cite{Bakolas2016Optimal}, the covariance steering problem is formulated as a convex program.
Additional state chance constraints are considered in \cite{Okamoto2018Optimal} and nonlinear systems are considered in \cite{Ridderhof2019nonlinear} using iterative linearization.
The covariance steering problem with output feedback has also been studied in \cite{Ridderhof2020Chance,Chen2015Steering,Bakolas2017Covarince}.

In this paper, we propose the CS-BRM algorithm, 
which uses covariance steering as the edge controller of a BRM to ensure a priori node reachability.
Since the goal of covariance steering is to reach a given distribution of the state, it is well-suited for reaching a belief node. % and thus provides a way of addressing the ``curse of history'' problem.
In addition, covariance steering avoids the limitation of sampling in the equilibrium space,
and thus the proposed CS-BRM method allows sampling of non-stationary belief nodes. 
Our method allows searching in the velocity space and thus finds paths with lower cost, as demonstrated in the numerical examples in Section~\ref{Sec:Implementation}.

Extending belief planning to nonlinear systems requires a nominal trajectory for each edge \cite{Agha-Mohammadi2014FIRM}.
In the SLQG-FIRM framework,
these nominal trajectories are either assumed to be given or being approximated by simple straight lines \cite{Van2011LQG-MP, Agha-Mohammadi2014FIRM}.
However, the nominal trajectory has to be dynamically feasible in order to apply a time-varying LQG controller. 
Also, the nominal trajectory must be optimal if one wishes to generate optimal motion plans.
Finding an \textit{optimal} nominal trajectory requires solving a two-point boundary value problem (TPBVP), which 
for general nonlinear systems can be computationally 
expensive~\cite{Betts1998Survey}.
In this paper, we develop a simple, yet efficient, algorithm to find suitable nominal trajectories between the nodes of the BRM graph to steer the mean states in an optimal fashion.

The contributions of the paper are summarized as follows.

\begin{itemize}

\item[$\bullet$] A new belief space planning method, called CS-BRM, is developed, to construct a roadmap in belief space by using the recently developed theory of
finite-time covariance control.
CS-BRM achieves finite-time belief node reachability using covariance steering %addresses the ``curse of history'' problem 
and overcomes the limitation of sampling stationary nodes.
%Compared to previous BRM methods, CS-BRM allows searching the velocity space of the belief nodes and it does not require a convergence phase at every node.

\item[$\bullet$] The concept of \textit{compatible nominal trajectory} is introduced, which aims to improve the performance of linearization-based control methods to control nonlinear systems.
An efficient algorithm, called the CNT algorithm, is proposed to compute nominal feasible trajectories for nonlinear systems.
%Instead of relying on complicated numerical solvers to solve the TPBVP, we develop an algorithm that utilizes the analytical solution of the mean control of a linear time-varying system along with iterative linearizations.
%We show that the proposed algorithm can find nominal trajectories for nonlinear systems quickly.

\end{itemize}
 
%The method deals with the covariance steering problem for nonlinear systems with output feedback.
%It is based on the separation of observation and control, separation of mean control and covariance control, and the CNT algorithm.
%We compare CS-BRM with the current state-of-the-art SLQG-FIRM method and show that by sampling non-stationary belief nodes and searching the velocity space, our method is able to find more efficient and lower-cost plans.

The paper is organized as follows.
The statement of the problem is given in Section \ref{Sec:ProblemStatement}.
In Section~\ref{Sec:CovSteering}, the output-feedback covariance steering theory for linear systems is outlined. 
%The covariance steering controller is used as the edge controller of the BRM edges.
In Section \ref{Sec:nonlinearCS}, the method of computing nominal trajectory for nonlinear systems is introduced.
The main algorithm, CS-BRM, is given in Section \ref{Sec:CS-BRM}.
The numerical implementation of the proposed algorithm is presented in 
Section~\ref{Sec:Implementation}.
Finally, Section \ref{Sec:Conclusion} concludes the paper.

\section{Problem Statement} \label{Sec:ProblemStatement}

We consider the problem of planning for a nonholonomic robot in an uncertain environment which contains obstacles.
The uncertainty in the problem stems from model uncertainty, as well from sensor noise that corrupts the measurements.
We model such a system by a  stochastic difference equation of the form
\begin{equation}   \label{eq:nonlinear}
		x_{k+1} = f(x_k,u_k,w_k),
\end{equation}
where $k=0,1,\ldots,N-1$ are the discrete time-steps, $x_k \in \mathbb{R}^{n_x}$ is the state, and $u_k \in \mathbb{R}^{n_u}$ is the control input. 
The steps of the noise process $w_k \in \mathbb{R}^{n_w}$ are i.i.d standard Gaussian random vectors. 
The measurements are given by the noisy and partial sensing model
\begin{equation}   \label{eq:nonlinear:sensor}
	y_k = h(x_k,v_k),
\end{equation}
where $y_k \in \mathbb{R}^{n_y}$ is the measurement at time step $k$, and the steps of the process $v_k \in \mathbb{R}^{n_y}$ are i.i.d standard Gaussian random vectors.
We assume that $(w_k)_{k=0}^{N-1}$ and $(v_k)_{k=0}^{N-1}$ are independent.

The objective is to steer the system (\ref{eq:nonlinear}) from some initial state $x_0$ to some final state $x_{N}$  within $N$ time steps while avoiding obstacles and, at the same time, minimize a given performance index.
The controller  $u_k$ at time step $k$ is allowed to depend on the whole history of measurements $y_\ell,~\ell = 0,\ldots,k$ up to time $k$ but not on any future measurements.

%The edges of BRM are obtained by using edge controllers that steer the robot from one belief to another to obtain
%a directed graph.
%After constructing the BRM, the path is found using a graph search algorithm.
%However, the construction of the edge controllers for BRMs is not trivial, as it is n
% The main remaining pieces for constructing the belief space roadmap are to explicitly define the belief node, construct the edge controller, and define the edge cost.
% (\ref{edgecontrollereq1}) and (\ref{edgecontrollereq2}) may represent the linearized model of the nonlinear system along a nominal trajectory.

This is a difficult problem to solve in its full generality.
Here we use graph-based methods to build a roadmap in the space of distributions of the states (e.g., the belief space) for multi-query motion planning.
Planning in the belief space accounts for the uncertainty inherent in system (\ref{eq:nonlinear}) and (\ref{eq:nonlinear:sensor}) owing to the noises $w_k$ and $v_k$, as well as uncertainty owing to the 
distribution of the initial states.
Each node sampled in the belief space is a distribution over the state. 
The edges in a BRM steer the state from one distribution to another. 
In the next section, we describe a methodology to design BRM edge controllers that allow to steer from one node (i.e., distribution) to another. 
Similar to previous works \cite{Prentice2009The, Agha-Mohammadi2014FIRM, Van2011LQG-MP}, we consider Gaussian distributions where the belief is given by the state mean and the state covariance.

\section{Covariance Steering} \label{Sec:CovSteering}

In this section, we give a introduction of covariance steering and outline some key results.
Some of these results are explicitly used in subsequent sections.
The derivation in this section follows closely~\cite{Ridderhof2020Chance, Okamoto2018Optimal}.
% The difference from~\cite{Ridderhof2020Chance} is that we will separate the covariance control and mean control.
%Also, we consider a drift term in the system model and a reference trajectory in the cost function. % which are useful for our problem formulation.
%Then, we will introduce our method to compute nominal trajectories of system (\ref{eq:nonlinear})-(\ref{eq:nonlinear:sensor}).

% It is assumed that the nonlinear system (\ref{eq:nonlinear})-(\ref{eq:nonlinear:sensor}) 
% can be well approximated locally by its linearization about a nominal trajectory.
Given a nominal 
trajectory $(n_k)_{k=0}^{N-1}$, where $ n_k = (x_k^r, u_k^r)$,
we can construct a linear approximation of (\ref{eq:nonlinear})-(\ref{eq:nonlinear:sensor}) around $(n_k)_{k=0}^{N-1}$ via linearization as follows
%Note that we do not need to linearized the system at the last time step. 
%Recall that we also define a reference state trajectory $(m_k)_{k=0}^{N-1}$ in Problem 1.  
%The nominal state trajectory $(x_k^r)_{k=0}^{N-1}$ and $(m_k)_{k=0}^{N-1}$ are not necessarily the same. 
%The trajectory $(m_k)_{k=0}^{N-1}$ affects the cost function, namely, we want the state trajectory to stay close to $(m_k)_{k=0}^{N-1}$.
%The nominal trajectory $(n_k)_{k=0}^{N-1}$ is the trajectory along which the nonlinear system linearization is performed.
\begin{align}    	
	x_{k+1} &= A_k x_{k} + B_k u_k + h_k + G_k w_k,  \label{edgecontrollereq1} \\
		y_k &=C_k x_k + D_k v_k,  \label{edgecontrollereq2}
\end{align}
where
 $h_k \in \mathbb{R}^{n_x}$ is the drift term, $A_k \in \mathbb{R}^{n_x \times n_x}$, $B_k \in \mathbb{R}^{n_x \times n_u}$, and $G_k \in \mathbb{R}^{n_x \times n_x}$ are system matrices,
 and $C_k \in \mathbb{R}^{n_y \times n_x}$ and $D_k \in \mathbb{R}^{n_y \times n_y}$ are observation model matrices.

We define the covariance steering problem as follows.

\textit{Problem 1:} Find the control sequence $u=(u_k)_{k=0}^{N-1}$ such that the system given by (\ref{edgecontrollereq1}) and (\ref{edgecontrollereq2}), starting from the initial state distribution $x_0 \sim \mathcal{N}(\bar{x}_0,P_0)$, reaches the final distribution $x_N \sim \mathcal{N}(\bar{x}_N,P_N)$ where
$P_N \preceq P_f$,
while minimizing the cost functional
\begin{equation}
    J(u) = \mathbb{E} \left[ \sum_{k=0}^{N-1} (x_k - m_k)^\top Q_k (x_k - m_k) + u_k^\top R_k u_k \right],
    \label{edgecontrollereq3}
\end{equation}
where $(m_k)_{k=0}^{N-1}$ is a given reference trajectory of the states, $\bar{x}_k = \mathbb{E} (x_k)$ is the mean of the state $x_k$, $P_0$ and $P_N$ are the covariance matrices of $x_0$ and $x_N$, respectively, the terminal covariance $P_N$ is upper bounded by a given covariance matrix $P_f$, and the matrices $(Q_k \succeq 0)$ and $(R_k \succ 0)$ are given.
%problem parameters.

\subsection{Separation of Observation and Control}

Similarly to \cite{Ridderhof2020Chance}, we assume that the control input $u_k$ at time-step $k$ is an affine function of the measurement data.
%This restriction is made to ensure that if $x_k$ is Gaussian, then $x_{k+1}$ will also be Gaussian.
It follows that the state will be Gaussian distributed over the entire horizon of the problem. 
We assume the system (\ref{edgecontrollereq1})-(\ref{edgecontrollereq2}) is observable.
To solve Problem 1, we use a Kalman filter to estimate the state.
Specifically, let the prior initial state estimate be $\hat{x}_{0\m}$ and distributed according to $\hat{x}_{0\m} \sim \mathcal{N}(\bar{x}_0, \hat{P}_{0\m})$, and let
 the prior initial estimation error be $\tilde{x}_{0\m} = x_0 - \hat{x}_{0\m}$ and let its distribution be given by $\tilde{x}_{0\m} \sim \mathcal{N}(0, \tilde{P}_{0\m})$.
The estimated state at time $k$, denoted as $\hat{x}_k = \mathbb{E}[x_k | Y_k]$,
with $Y_k$ denoting the filtration generated by $\{\hat{x}_{0\m}, y_i : 0 \leq i \leq k)$,
is computed from~\cite{Bryson1975Applied}
\begin{equation}
\begin{split}
    \hat{x}_{k} & = \hat{x}_{k\m} + L_k (y_k - C_k \hat{x}_{k\m}), \\
    \hat{x}_{k\m} & = A_{k-1} \hat{x}_{k-1} + B_{k-1} u_{k-1} + h_{k-1},
    \label{KFdynamics}
\end{split}
\end{equation}
where 
\begin{equation}
\begin{split}
L_k & =\tilde{P}_{k\m} C_k^\top ( C_k \tilde{P}_{k\m} C_k^\top + D_k D_k^\top )^{-1}, \\
\tilde{P}_k & =( I - L_k C_k) \tilde{P}_{k\m}, \\
\tilde{P}_{k\m} & = A_{k-1} \tilde{P}_{k-1} A_{k-1}^\top +G_{k-1} G_{k-1}^\top,
\label{KFupdate}
\end{split}
\end{equation}
where $L_k$ is the Kalman gain and where 
%$\tilde{P}_k$ is the estimation error covariance. 
the covariances of $x_k$, $\hat{x}_k$ and $\tilde{x}_k$ are denoted as 
$P_k = \mathbb{E}[(x_k - \bar{x}_k) (x_k - \bar{x}_k)^\top]$, $\hat{P}_k = \mathbb{E}[(\hat{x}_k - \bar{x}_k) (\hat{x}_k - \bar{x}_k)^\top]$ and $\tilde{P}_k = \mathbb{E}[(x_k - \hat{x}_k)(x_k - \hat{x}_k)^\top]$, respectively. 
% From (\ref{KFupdate}), the evolution of $\tilde{P}_{k\m}$ and $\tilde{P}_k$ do not depend on the control $(u_k)_{k=0}^{N-1}$. 

% Using (\ref{S3Eq5}), we have
% \begin{equation}
% \mathbb{E} [x_k^\top Q_k x_k] = \mathrm{trace} (\tilde{P}_k Q_k) + \mathbb{E} [\hat{x}_k^\top Q_k \hat{x}_k].
% \end{equation}
It can be shown that
the cost functional (\ref{edgecontrollereq3}) can be written as
\begin{align}
    J(u) = & \mathbb{E} \left[ \sum_{k=0}^{N-1} \hat{x}_k^\top Q_k \hat{x}_k + u_k^\top R_k u_k \right] - 2  \sum_{k=0}^{N-1} \bar{x}_k^\top Q_k m_k \nonumber \\
    & + \sum_{k=0}^{N-1}(\mathrm{trace} (\tilde{P}_k Q_k) + m_k^\top Q_k m_k),
    \label{CostEq9}
\end{align}
where the last summation is deterministic and does not depend on the control and thus it can be discarded.

By defining the \textit{innovation process} $(\xi_k)_{k=0}^{N}$ as
\begin{equation}
    \xi_k = y_k - \mathbb{E} [y_k | Y_{k-1}],
    \label{innovp1}
\end{equation}
and noting that $\mathbb{E} [y_k | Y_{k-1}] = \mathbb{E}[C_k x_k + D_k v_k | Y_{k-1}] = C_k \hat{x}_{k\m}$, the estimated state dynamics in equation (\ref{KFdynamics}) can be rewritten as
\begin{equation}
    \hat{x}_{k+1} = A_k \hat{x}_k + B_k u_k + h_k + L_{k+1} \xi_{k+1},
    \label{Eqxhat}
\end{equation}
with $\hat{x}_0 = \hat{x}_{0\m} + L_0 \xi_0$.
We can then restate Problem 1 as follows.

\textit{Problem 2:} Find the control sequence $(u_k)_{k=0}^{N-1}$, such that the system (\ref{Eqxhat}) starting from the initial distribution $\hat{x}_{0\m} \sim \mathcal{N}(\bar{x}_0,P_0-\tilde{P}_{0\m})$ reaches the final distribution $\hat{x}_N \sim \mathcal{N}(\bar{x}_N,P_N-\tilde{P}_N)$, where
$P_N \preceq P_f$, while minimizing the cost functional
\begin{equation}
    \hat{J}(u) = \mathbb{E} \left[ \sum_{k=0}^{N-1} \hat{x}_k^\top Q_k \hat{x}_k + u_k^\top R_k u_k \right] - 2  \sum_{k=0}^{N-1} \bar{x}_k^\top Q_k m_k.
    \label{cost1}
\end{equation}

To summarize, the covariance steering problem of the state $x_k$ with output feedback has been transformed to a covariance steering problem of the estimated state $\hat{x}_k$, where $\hat{x}_k$ is computed using the Kalman filter. 
This transformation relies on the separation principle of estimation and control.
The solution to this problem is given in the next section.
%While Problem 1 is defined with respect to the unknown state $x_k$, Problem 2 is defined with respect to the known state estimate $\hat{x}_k$.
%Also, the noise term $G_k w_k$ in (\ref{edgecontrollereq1}) is replaced by the noise term $L_{k+1} \xi_{k+1}$ in (\ref{Eqxhat}). 

% Remark: If $P_f \leq \tilde{P}_N$, then the covariance steering problem is infeasible.

\subsection{Separation of Mean Control and Covariance Control}

%Problem 2 defines a covariance steering problem for the estimated state $\hat{x}_k$, which is Gaussian distributed.
In this section, we will separate Problem 2 into a mean control problem and a covariance control problem.

By defining the augmented vectors
$U_k  = [u_0^\top \ u_1^\top \ \cdots \ u_k^\top]^\top,$
$\Xi_k  = [\xi_0^\top \ \xi_1^\top \ \cdots \ \xi_k^\top]^\top,$ and
$\hat{X}_k = [\hat{x}_0^\top \ \hat{x}_1^\top \ \cdots \ \hat{x}_k^\top]^\top.$
We can compute $\hat{X}_k$ as follows
\begin{equation}
    \hat{X} = A \hat{x}_{0\m} + B U + H + L \Xi,
    \label{hatX}
\end{equation}
where $\hat{X} = \hat{X}_N$, $U = U_{N-1}$, $\Xi = \Xi_{N}$, and $A$, $B$, $H$, and $L$ are block matrices constructed using the system matrices in~(\ref{Eqxhat}).
Defining $\check{x}_{0\m} \triangleq \hat{x}_{0\m} - \bar{x}_0$, $\bar{X}\triangleq \mathbb{E}[\hat{X}]$, $\bar{U}\triangleq \mathbb{E}[U]$, $\check{X}\triangleq \hat{X} - \bar{X}$, and $\tilde{U} \triangleq U - \bar{U}$,
and using~(\ref{hatX}), it follows that
\begin{align}
    \bar{X} &= A \bar{x}_0 + B \bar{U} + H, \label{Xbar}  \\
    \check{X} &= A \check{x}_{0\m} + B \tilde{U} + L \Xi. \label{Xcheck}
\end{align}
The cost functional in (\ref{cost1}) can be rewritten as 
\begin{equation}
\begin{split}
    \hat{J}(u) = 
    %&\mathbb{E} \left[ \hat{X}^\top Q \hat{X} + U^\top R U \right] - 2 \bar{X}^\top Q M_r\\
    & \mathbb{E} [\check{X}^\top Q \check{X} + \tilde{U}^\top R \tilde{U}] + \bar{X}^\top Q \bar{X} + \bar{U}^\top R \bar{U} \\
    & - 2 \bar{X}^\top Q M_r,
    \label{S3Eq26}
\end{split}
\end{equation}
where $Q = \text{blkdiag}(Q_0,Q_1, \ldots ,Q_{N-1},0)$, $R = \text{blkdiag}(R_0,R_1, \ldots ,R_{N-1})$, $M_r = [m_0^\top \ m_1^\top \ \cdots \ m_N^\top]^\top$.

From (\ref{Xbar}), (\ref{Xcheck}), and (\ref{S3Eq26}), the covariance steering problem of the estimated state $\hat{x}$ (Problem 2) can thus be divided into the following mean control and covariance control problems.
The mean control problem is given by
\begin{equation}
\begin{split}
    \min_{\bar{U}} & \ \ \bar{X}^\top Q \bar{X} + \bar{U}^\top R \bar{U} - 2 \bar{X}^\top Q M_r \\
    \mathrm{s.t.} & \ \ \bar{X} = A \bar{x}_0 + B \bar{U} + H, \\
    & \ \ E_0 \bar{X}=\bar{x}_0, \quad E_N \bar{X}=\bar{x}_N,
    \label{MeanControl}
\end{split}
\end{equation}
where $E_k$ is a matrix defined such that $E_k \bar{X} = \bar{x}_k$.
The covariance control problem is given by
\begin{equation}
\begin{split}
    \min_{\tilde{U}} & \ \ \mathbb{E} [\check{X}^\top Q \check{X} + \tilde{U}^\top R \tilde{U}] \\
    \mathrm{s.t.} & \ \ \check{X} = A \check{x}_{0\m} + B \tilde{U} + L\Xi,  \\
    & \ \ \mathbb{E}[\check{x}_{0\m} \check{x}_{0\m}^\top] = \hat{P}_{0\m}, \ \ E_N \mathbb{E}[\check{X} \check{X}^\top] E_N^\top = \hat{P}_N,
    \label{CovControl}
\end{split}
\end{equation}
where $\hat{P}_{0\m} = P_0-\tilde{P}_{0\m}$, $\hat{P}_N = P_N-\tilde{P}_N \preceq P_f - \tilde{P}_N$.
%and the terminal equality constraint is relaxed to an inequality constraint.
% Note that $\check{P}_{0\m} = \hat{P}_{0\m}$ and $\check{P}_N = \hat{P}_N$.

\subsection{Solutions of the Mean Control and Covariance Control Problems} \label{MeanCovSol}

Assuming that the system (\ref{edgecontrollereq1}) is controllable,
the solution for the mean control problem (\ref{MeanControl}) to
obtain the mean trajectory $(\bar{x}_k)_{k=0}^{N}$ and $(\bar{u}_k)_{k=0}^{N-1}$
is given by~\cite{Okamoto2018Optimal}
\begin{equation}
\begin{split}
    \bar{U}^* = & W(- V + \bar{B}_N^\top(\bar{B}_N W \bar{B}_N^\top)^{-1} \\
    & (\bar{x}_N - \bar{A}_N \bar{x}_0 - H_N + \bar{B}_N W V)),
    \label{S3Eq29}
\end{split}
\end{equation}
where $W = (B^\top Q B + R)^{-1}$, and $V = B^\top Q (A \bar{x}_0 + H - M_r)$.

The solution to the covariance control problem (\ref{CovControl}) is given by the controller of the form
\begin{equation}
    \tilde{u}_k = \sum_{i=0}^{k} K_{k,i} \check{x}_i,
    \label{Eq38}
\end{equation}
equivalently, by $\tilde{U}=K \check{X}$, where 
the control gain matrix $K$ is lower block diagonal of the form
\begin{equation}
    K = \begin{bmatrix} K_{0,0} & 0 & \cdots & 0 & 0\\ K_{1,0} & K_{1,1} & \cdots & 0 & 0 \\ \vdots & \vdots & \ddots & \vdots & 0\\ K_{N-1,0} & K_{N-1,1} & \cdots & K_{N-1,N-1} & 0 \end{bmatrix},
\end{equation}
and is computed from $K = F(I + BF)^{-1}$ where $F$
is obtained by solving the following convex 
optimization problem~\cite{Ridderhof2020Chance}
\begin{align}
   \min_F \ \mathrm{trace} [((I+BF)^\top Q(I+BF) + F^\top R F) P_Z],   \label{CovCost}
\end{align}  
subject to
\begin{equation}
    \lVert P_Z^{1/2}(I+BF)^\top E_N^\top (P_f - \tilde{P}_N)^{-1/2} \rVert - 1 \leq 0,
    \label{CovConstraint}
\end{equation}
where $P_Z = A\hat{P}_{0\m} A^\top + LP_{\Xi} L^\top$ is the covariance of $Z \triangleq A \check{x}_{0\m} + L \Xi$  and
$P_{\Xi} = \text{blkdiag}(P_{\xi_0},\cdots,P_{\xi_N})$ is the covariance of the innovation process.

\section{Computing the nominal Trajectory} \label{Sec:nonlinearCS}

%The covariance steering problem for stochastic, discrete, linear time-varying systems was studied in  Section~\ref{Sec:CovSteering}. 
In this section, we develop an efficient algorithm to find nominal trajectories for nonlinear systems using the mean controller (\ref{S3Eq29}).
% Consider again the  nonlinear stochastic system with dynamics  (\ref{eq:nonlinear}). 
% The system is linearized along a nominal trajectory $(n_k)_{k=0}^{N-1}$ to obtain 
% the system shown in (\ref{edgecontrollereq1})-(\ref{edgecontrollereq2}).
%Note that we do not need to linearized the system at the last time step. 
%Recall that we also define a reference state trajectory $(m_k)_{k=0}^{N-1}$ in Problem 1.  
%The nominal state trajectory $(x_k^r)_{k=0}^{N-1}$ and $(m_k)_{k=0}^{N-1}$ are not necessarily the same. 
%The trajectory $(m_k)_{k=0}^{N-1}$ affects the cost function, namely, we want the state trajectory to stay close to $(m_k)_{k=0}^{N-1}$.
%The nominal trajectory $(n_k)_{k=0}^{N-1}$ is the trajectory along which the nonlinear system linearization is performed.
%Given the nominal trajectory, we obtain a discrete, linear time-varying model of the form (\ref{edgecontrollereq1}) and (\ref{edgecontrollereq2}) by linearization, where
%the system matrices are
%\begin{equation}
%\begin{split}
%    A_k &= \frac{\partial f}{\partial x}|_{(x_k^r,u_k^r,0)}, \quad B_k = \frac{\partial f}{\partial u}|_{(x_k^r,u_k^r,0)}, \\
%    G_k &= \frac{\partial f}{\partial w}|_{(x_k^r,u_k^r,0)}, \quad h_k = f(x_k^r,u_k^r,0) - A_k x_k^r -B_k u_k^r, \\
%    C_k &= \frac{\partial h}{\partial x}|_{(x_k^r,0)}, \quad D_k = \frac{\partial h}{\partial v}|_{(x_k^r,0)}, \\
%    g_k &= h(x_k^r,0) - C_k x_k^r,
%    \label{systemMatrices}
%\end{split}
%\end{equation}
% Finding the nominal trajectory $(n_k)_{k=0}^{N-1}$ itself is an open-loop controller design problem. 
A deterministic nonlinear dynamic model, i.e., with the noise $w_k$ set to zero, is considered when designing the nominal trajectory.
% The nominal trajectory is typically obtained by solving a TPBVP considering the deterministic dynamics. %that satisfies the deterministic dynamics and steers the system from an initial state to a goal state.
% Next, we define a class of nominal trajectories, referred to as \textit{compatible nominal trajectories}, and we develop an algorithm to find them. 
%The compatible nominal trajectory is introduced for the control of nonlinear systems using linerization-based methods.
To this end, consider the deterministic nonlinear system 
\begin{equation}   \label{eq:nomsystem} 
x_{k+1} = f(x_k,u_k,0).
\end{equation}
Let $(x_k^r)_{k = 0}^{N-1}$ and $(u_k^r)_{k = 0}^{N-1}$ be a
nominal trajectory for this system.
%The nominal trajectory may not be dynamically feasible. That is, the control sequence and state sequence may not satisfy the dynamics $x_{k+1} = f(x_k,u_k,0)$.
%By linearizing the system along this nominal trajectory, we obtain a discrete-time linear time-varying model. 
%This discrete-time linear time-varying model is a local approximation of the nonlinear system and is used for trajectory generation.
The linearized model (\ref{edgecontrollereq1}) along this nominal trajectory
will be used by the mean controller (\ref{S3Eq29}) to compute a control sequence $(u_k^c)_{k = 0}^{N-1}$ and the corresponding state sequence $(x_k^c)_{k = 0}^{N-1}$. 
%That is, $u_k^c$  and $x_k^c$ satisfy the mean dynamics (\ref{Xbar}).
% Next, we give the definition of a compatible
% nominal trajectory

\textit{Definition 1:} A nominal trajectory is called a \textit{compatible nominal trajectory} for system (\ref{eq:nomsystem}) 
if $x_k^r = x_k^c$ and $u_k^r = u_k^c$ for $k=0, \dots, N-1$.
%Consider the control of a deterministic nonlinear system using the linearization method, a nominal trajectory is called a \textit{Compatible nominal trajectory} if the resulting trajectory of the linearized system using the designed control coincide with the nominal trajectory.

Linearizing along a compatible nominal trajectory ensures that the nonlinear system is linearized at the ``correct'' points.
When applying the designed control to the linearized system, the resulting trajectory is the same as the nominal trajectory, which means that the system reaches the exact points where the linearization is performed.
On the other hand, there will be extra errors caused by the linearization if the system is linearized along a non-compatible nominal trajectory.

The iterative algorithm to find a compatible nominal trajectory is given in Algorithm \ref{alg:CNT}.

% First, we initialize the algorithm with an initial guess for the nominal trajectory. 
% %The initial guess does not have to be dynamically feasible for (\ref{eq:nomsystem}). For example, one may choose a straight line at each state and control dimension as the initial guess.
% Next, we linearize the nonlinear system (\ref{eq:nomsystem}) along this nominal trajectory. 
% By solving the mean control problem (\ref{MeanControl}) using (\ref{S3Eq29}), we obtain the mean trajectory.
% Then, we update the nominal trajectory using the mean trajectory. 
% We repeat the above three steps until converge or a stop criterion is satisfied.
%A commonly used convergence criterion is to check if the change of $(x_k^r)_{k = 0}^{N-1}$ between successive iterations is smaller than a small threshold. 

%%%%%%%%%%%%%%%%%%%%%%%%%%%%%%%%%% Compatible Nominal Trajectory %%%%%%%%%%%%%%%%%%%%%%%%%%%%%%%%%%%%
\IncMargin{.5em}
\begin{algorithm}
\caption{Compatible Nominal Trajectory (CNT)}
\label{alg:CNT}
Initialize $(x_k^r)_{k = 0}^{N-1}$ and $(u_k^r)_{k = 0}^{N-1}$ \;
\While{NotConverged} 
{   Linearize (\ref{eq:nomsystem}) along $(x_k^r)_{k = 0}^{N-1}$ and $(u_k^r)_{k = 0}^{N-1}$ \;
    Compute the mean optimal control using (\ref{S3Eq29}) to obtain the control sequence $(u_k^c)_{k = 0}^{N-1}$ \;
    Compute the the controlled state trajectory $(x_k^c)_{k = 0}^{N-1}$ using (\ref{Xbar}) \;
    Set $(x_k^r = x_k^c)_{k = 0}^{N-1}$ and $(u_k^r = u_k^c)_{k = 0}^{N-1}$ \;
}
\KwRet $(x_k^r)_{k = 0}^{N-1}$ and $(u_k^r)_{k = 0}^{N-1}$;
\end{algorithm}
\DecMargin{.5em}
%%%%%%%%%%%%%%%%%%%%%%%%%%%%%%%%%%%%%%%%%%%%%%%%%%%%%%%%%%%%%%%%%%%%%%%%%%%%%%%%%%%%

When the algorithm converges, the nominal trajectory is the same as the mean optimal trajectory, which, by definition, is a compatible nominal trajectory.
Note that $(\ref{S3Eq29})$ gives the analytical solution of the mean control, which can be quickly computed. 
Each iteration of the algorithm has a low computational load and the overall algorithm may be solved efficiently. 
As seen later in Section~\ref{Sec:Implementation}, the algorithm typically convergences within a few iterations.

% Appendix~\ref{AppendixForProp2} shows how to choose the reference state trajectory $(m_k)_{k=0}^{N-1}$ in (\ref{edgecontrollereq3}).
% Choosing the reference state trajectory $(m_k)_{k=0}^{N-1}$ judiciously has the benefit of achieving a lower cost, and increasing the stability of the Algorithm~\ref{alg:CNT}.
% Note that the cost functional $(\ref{edgecontrollereq3})$ regulates the state trajectories to stay close to $(m_k)_{k=0}^{N-1}$.
% By setting $m_k = x_k^r$ for all $k = 0, \ldots, N-1$, the resulting state trajectory is regulated to be close to the nominal trajectory, and thus the linearization of the nonlinear system is more likely to be locally valid, which will help the stability of Algorithm \ref{alg:CNT}. %to find a compatible nominal trajectory.

\begin{figure}[htb]
    \centering
    \begin{subfigure}[b]{0.48\columnwidth}
         \centering
         \includegraphics[width=1\columnwidth]{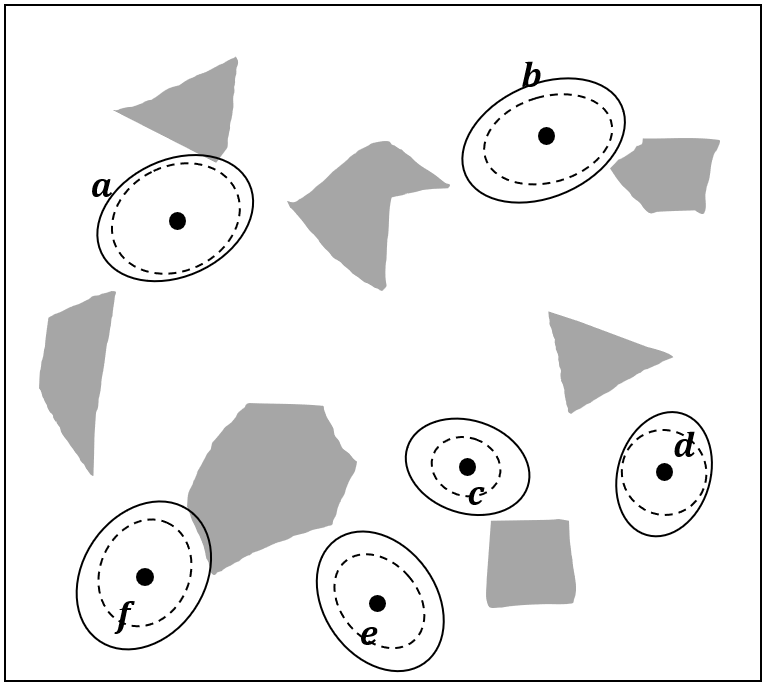}
         \caption{}
     \end{subfigure}
     \begin{subfigure}[b]{0.48\columnwidth}
         \centering
         \includegraphics[width=1\columnwidth]{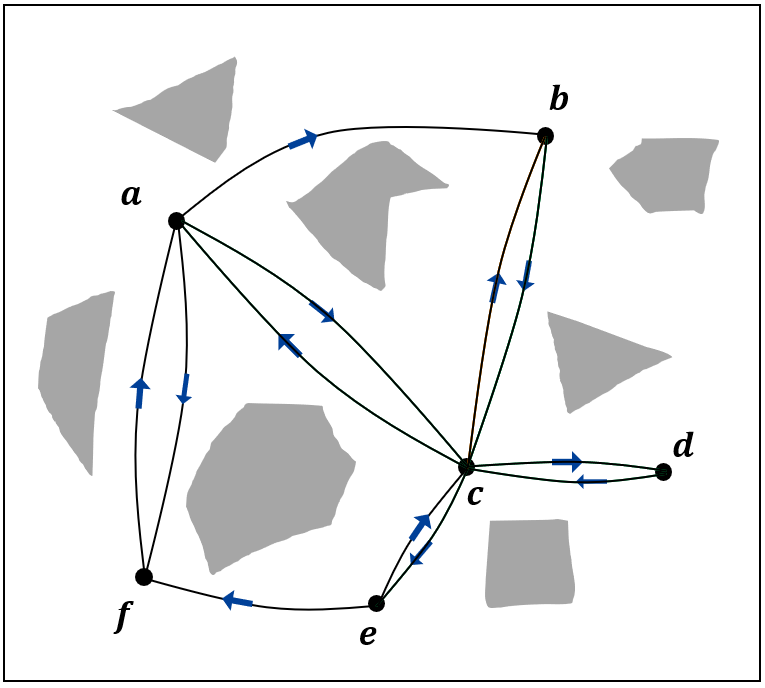}
         \caption{}
     \end{subfigure} \\
     \begin{subfigure}[b]{0.48\columnwidth}
         \centering
         \includegraphics[width=1\columnwidth]{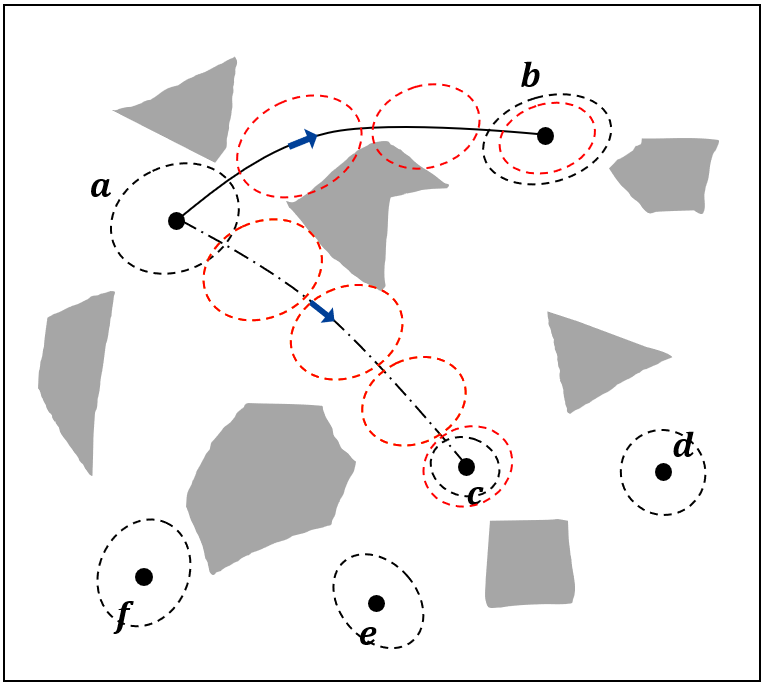}
         \caption{}
     \end{subfigure}
     \begin{subfigure}[b]{0.48\columnwidth}
         \centering
         \includegraphics[width=1\columnwidth]{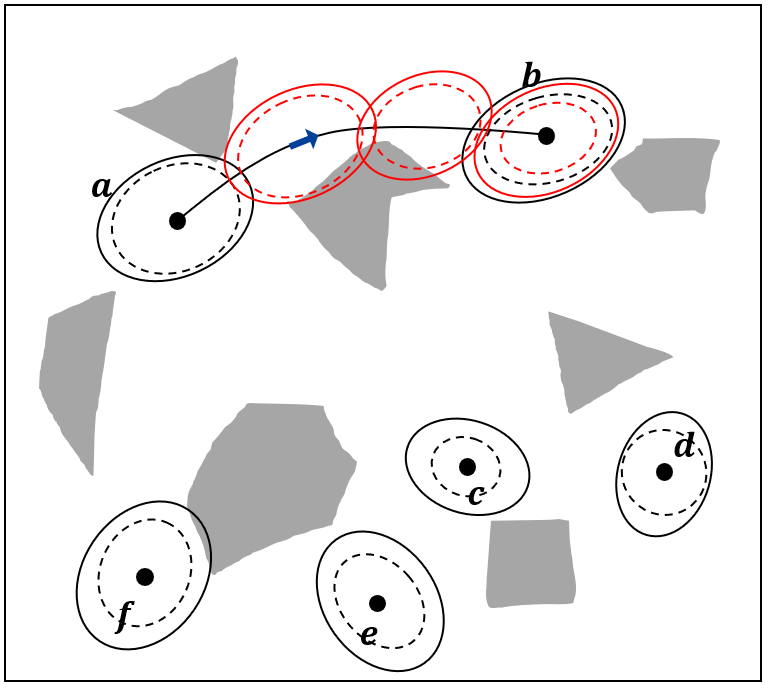}
         \caption{}
     \end{subfigure}
        \caption{Construction of the CS-BRM. 
        	Gray shapes denote obstacles. 
        	(a) Node sampling. Each node is represented by its mean, a state covariance, and a state estimation error covariance, all of which are sampled in proper sample spaces; 
        	(b) Mean trajectories are computed for all neighboring node pairs using the mean controller. 
        	Only mean trajectories that are collision-free are preserved. 
        	Those mean trajectories indicate possible connections/edges in the CS-BRM; 
        	(c) Kalman filter updates are simulated for each possible edge; % which gives the state estimation error covariance at every time step along that edge; 
        	(d) Covariance control is applied for each edge to execute the transition between the nodes.}
        \label{PRMProce}
\end{figure}

\section{The CS-BRM Algorithm} \label{Sec:CS-BRM}

The nodes in the CS-BRM are sampled in the belief space. 
In a partially observable environment, the belief $b_k$ at time-step $k$  is given by the conditional probability distribution of the state $x_k$, conditioned on the history of observations  $(y_i)_{i=0}^{k}$ and the history of control inputs $(u_i)_{i=0}^{k-1}$, that is, $b_k = \mathbb{P}(x_k|(y_i)_{i=0}^{k},(u_i)_{i=0}^{k-1})$. %\mathrm{Pr}
In a Gaussian belief space, $b_k$ can be equivalently represented by the estimated state, $\hat{x}_k$, and the estimation error covariance, $\tilde{P}_k$, that is, $b_k = (\hat{x}_k, \tilde{P}_k)$ \cite{Bry2011rapidly,Agha-Mohammadi2014FIRM}.
The state estimate $\hat{x}_k$ is Guassian, and is given by $\hat{x}_k \sim \mathcal{N}(\bar{x}_k, \hat{P}_k)$.
Hence, the Gaussian belief can be also written as $b_k = (\bar{x}_k, \hat{P}_k, \tilde{P}_k)$.
%Similar to the traditional PRM, the probabilistic roadmap in belief space is a graph consisting of nodes and edges. Note that the graph is directed.
The main idea of CS-BRM is to use covariance steering theory to design the edge controller to achieve node reachability.
%%%%%%%%%%%%%%%%%%%%%%%%%%%%%%%%%%%%%%%%%%%%%%%%%%%%%%%%%%%%%%%%%%%%%%%%%%%
\begin{algorithm}
\caption{Constructing CS-BRM}
\label{alg:CSBRM}
$V=\{nod_1, \ \dots, \ nod_n\} \leftarrow \texttt{SampleNodes}(n)$ \;
$V_r \leftarrow V$, $E \leftarrow \emptyset$ \;
\For{$i=1:n$} 
{
    $V_{near} \leftarrow \texttt{Neighbor}(V_r, \ nod_i)$\;
    \ForEach{$nod_j \in V_{near}$}
    {   
        $(\bar{U}_{ij}, \ \tau_{ij}, \ MCost_{ij}) \leftarrow  \texttt{MTraj}(nod_{i}, \ nod_j)$ \;
        \If {$\texttt{ObstacleFree}(\tau_{ij})$}
        {
            $\tilde{P}_{N\m} \leftarrow \texttt{KF}(nod_i, \ nod_j)$ \;
            \If {$\tilde{P}_{N\m} \preceq \tilde{P}_{nod_j}$}
            {
                $(\tilde{U}_{ij}, \ CovCost_{ij}) \leftarrow \texttt{CovControl}(nod_i, \ nod_j)$ \;
                $CollisionCost_{ij} \leftarrow \texttt{MonteCarlo}(\bar{U}_{ij}, \ \tilde{U}_{ij})$ \;
                $E_{ij} \leftarrow (\bar{U}_{ij}, \ \tilde{U}_{ij}, \ EdgeCost_{ij})$ \;
                $E \leftarrow E \cup E_{ij}$
            }
        }
        $(\bar{U}_{ji}, \ \tau_{ji}, \ MCost_{ji}) \leftarrow  \texttt{MTraj}(nod_{j}, \ nod_i)$ \;
        \If {$\texttt{ObstacleFree}(\tau_{ji})$}
        {
            Repeat line 8-13 with $i$ and $j$ swapped 
        }
    }
    $V_r \leftarrow V_r \setminus nod_i$ \;
}
$\text{CS-BRM} \leftarrow (V, \ E)$ \;
\KwRet $\text{CS-BRM}$;
\end{algorithm}
%%%%%%%%%%%%%%%%%%%%%%%%%%%%%%%%%%%%%%%%%%%%%%%%%%%%%%%%%%%%%%%%%%%%%%%%%%%

An illustration of the steps for building the CS-BRM is shown in Fig.~\ref{PRMProce}.
The algorithm for constructing the CS-BRM is given in Algorithm \ref{alg:CSBRM}.
The following procedures are used in the algorithm. \\
%%%%
\textbf{Sample Nodes:} The function \texttt{SampleNodes}$(n)$ samples $n$ CS-BRM nodes.
A node in the CS-BRM is represented by the tuple $(\bar{x}, \hat{P}_{\m}, \tilde{P}_{\m})$, where $\bar{x}$ is the state mean, $\hat{P}_{\m}$ is the prior estimated state covariance, and $\tilde{P}_{\m}$ is the prior state estimation error covariance.
Since $P = \hat{P}_{\m} + \tilde{P}_{\m}$, the node can also be equivalently represented by $(\bar{x}, P, \tilde{P}_{\m})$. Note that we can compute the a posteriori convariances $\hat{P}$ and $\tilde{P}$ using $\hat{P}_{\m}$ and $\tilde{P}_{\m}$. 
%Thus, we may also represent the node by $(\bar{x}, \hat{P}, \tilde{P})$.
For constructing node $j$, the algorithm starts by sampling the free state space, which provides the mean $\bar{x}_j$ of the distribution. 
Then, an estimated state covariance $\hat{P}_{j\m}$ is sampled from the space of symmetric and positive definite matrix space.
%For simplicity, in our implementation we consider diagonal matrices and sample positive real numbers for the diagonal entries.
Similarly, a state estimation error covariance $\tilde{P}_{j\m}$ is also sampled at node $j$. 
$\tilde{P}_{j\m}$ is the initial condition of the Kalman filter update for the edges that are coming out of node $j$.
In Fig.~\ref{PRMProce}(a), for each node, $\bar{x}$ is shown as a black dot, $P$ is shown as a solid ellipse, and $\tilde{P}_{\m}$ is shown as a dashed ellipse. \\ %%%%
\textbf{Neighbor:} The function \texttt{Neighbor}$(V_r, nod_i)$ finds all the nodes in $V_r$ that are within a given distance $d_1$ to node $nod_i$, where $V_r$ is a node set containing all nodes in the CS-BRM. \\
%Similar to the traditional PRM, every node tries to connect with other nodes in the graph that are within a distance to itself.
%One way to compute the distance between two nodes, $nod_i$ and $nod_j$, is to use their state means $\bar{x}_i$ and $\bar{x}_j$. We use the Euclidean distance in the position space as the distance metric.
%%%%
\textbf{Mean Trajectory:} Given two nodes $nod_i$ and $nod_j$, \texttt{MTraj}$(nod_i, nod_j)$ uses the mean controller (\ref{S3Eq29}) and Algorithm \ref{alg:CNT} to find the compatible nominal trajectory, which is also the mean trajectory from $nod_i$ to $nod_j$.
The function returns the mean control $\bar{U}_{ij}$, mean trajectory $\tau_{ij}$, and the cost of the mean control $MCost_{ij}$. \\
%%%%
\textbf{Obstacle Checking:} The function \texttt{ObstacleFree}$(\tau_{ij})$ returns true if the mean trajectory $\tau_{ij}$ is collision free. \\
%%%%
\textbf{Kalman Filter:} Given two nodes $nod_i$ and $nod_j$, \texttt{KF}$(nod_i, \ nod_j)$ returns the prior estimation error covariance at the last time step of the trajectory from $nod_i$ to $nod_j$. \\
%%%%
\textbf{Covariance Control:} \texttt{CovControl}$(nod_i, nod_j)$ solves the covariance control problem from $nod_i$ to $nod_j$.
It returns the control $\tilde{U}_{ij}$ and the cost $CovCost_{ij}$. \\
%%%%
\textbf{Monte Carlo:} We use Monte Carlo simulations to calculate the probability of collision of the edges.
For edge $E_{ij}$, the initial state $x_0$ and initial state estimate $\hat{x}_{0\m}$ are sampled from their corresponding distributions.
The state trajectory is simulated using the mean control $\bar{U}_{ij}$ and covariance control $\tilde{U}_{ij}$.
Then, collision checking is performed on the simulated state trajectory.
By repeating this process, we approximate the probability of collision of this edge.
In our implementation the collision cost $CollisionCost_{ij}$ is taken to be proportional to the probability of collision along that edge.
% Based on the probability of collision, a collision cost, $CollisionCost_{ij}$, is computed.

Algorithm \ref{alg:CSBRM} starts by sampling $n$ nodes in the belief space using \texttt{SampleNodes} (Line 1).
Lines 3-17 are the steps to add CS-BRM edges.
Given two neighboring nodes $nod_i$ and $nod_j$, Lines 6-13 try to construct the edge $E_{ij}$ and Lines 14-16 try to construct the edge $E_{ji}$.

Every edge in the CS-BRM is constructed by solving a covariance steering problem.
Each node tries to connect to its neighboring nodes if possible.
First, for each edge, we use Algorithm~\ref{alg:CNT} to find the mean trajectory of that edge (Line 6).
Next, we check if the mean trajectory is collision-free.
Only if the mean trajectory is collision-free do we proceed to the next step of edge construction.
%If the mean trajectory is in collision, the edge will not be added to the graph.
%
Then, the Kalman filter updates are simulated, which give the state estimation error covariance at every time step along that edge.
The state estimation error covariances of the edges $\overline{ab}$ and $\overline{ac}$ are shown as red dash ellipses in Figure~\ref{PRMProce}(c). 
%Take the edge $\overline{ab}$ as an example. If the state estimation error covariance at the final time step of $\overline{ab}$ is less than or equal to the state estimation error covariance at node $b$, the edge $\overline{ab}$ will be added to CS-BRM. 
%
Take the edge $\overline{ab}$ as an example. 
The initial condition of the Kalman filter is $\tilde{P}_{0\m} = \tilde{P}_{a\m}$.
The prior estimation error covariance at the final time-step, $\tilde{P}_{N\m}$, is compared with the state estimation error covariance of node $b$, $\tilde{P}_{b\m}$.
If $\tilde{P}_{N\m} \preceq \tilde{P}_{b\m}$ (Line 9), the algorithm proceeds to solve the covariance control problem and this edge is added to the graph.
For the covariance control problem of edge $\overline{ab}$, the initial constraint  and the terminal constraint are given by $\hat{P}_{0\m} = \hat{P}_{a\m}$ and %$\hat{P}_{N\m} \preceq \hat{P}_{b\m}$ or 
$\hat{P}_{N} \preceq \hat{P}_{b}$.
The covariance $\hat{P}_b$ is computed using $\hat{P}_{b\m}$ and $\tilde{P}_{b\m}$.
The final edge controller is the combination of the mean controller, covariance controller, and the Kalman filter.
% In Figure \ref{PRMProce}(c),~$\overline{ab}$ will be added as a CS-BRM edge, while $\overline{ac}$ will not. 

In addition to the edge controller, the edge cost is computed for each edge. 
%In order to provide more accurate and faster ways to compute collision probabilities, different methods have been developed \cite{Patil2012Estimating, Du2010Robotic, Agha-Mohammadi2014FIRM}.
%Monte Carlo simulations provide more accurate collision probabilities at the cost of expensive computation. Since the roadmap is constructed offline, this is not a problem for our method.
The edge cost $EdgeCost_{ij}$ is a weighted sum of $MCost_{ij}$, $CovCost_{ij}$, and $CollisionCost_{ij}$.
With covariance steering serving as the lower-level controller, the higher-level motion planning problem using the roadmap is a graph search problem similarly to a PRM. 
Thus, the covariance steering approach transforms the belief space roadmap into traditional PRM with specific edge costs.

In CS-BRM, each edge is an independent covariance steering problem and the planned path using CS-BRM consists of 
a concatenation of edges.
Since the terminal estimated state covariance satisfies an inequality constraint (\ref{CovControl}) and the terminal state estimation error covariance satisfies an inequality relation by construction (Line 9), it is important to verify that the covariance constraints at all nodes are still satisfied by concatenating the edges.

To this end, consider the covariance steering problem from node $a$ to node $b$. 
% Denote $\hat{P}_a$ and $\tilde{P}_{a}$ the estimated state covariance and state estimation error covariance at node $a$, $\hat{P}_b$ and $\tilde{P}_{b}$ the estimated state covariance and state estimation error covariance at node $b$. 
For the edge $\overline{ab}$, we have the initial constraints $\hat{P}_{0\m} = \hat{P}_{a\m}$, $\tilde{P}_{0\m} = \tilde{P}_{a\m}$, and the terminal covariance constraint 
%$\hat{P}_{N\m} \preceq \hat{P}_{b\m}$ and
$\hat{P}_{N} \preceq \hat{P}_b$.
Note that the constraint $\tilde{P}_{N\m} \preceq \tilde{P}_{b\m}$
is satisfied for $\overline{ab}$ (Line 9, Algorithm \ref{alg:CSBRM}).
Suppose that the solution of the covariance control problem of edge $\overline{ab}$ is the feedback gain $K$. 
By concatenating the edges, the system may not start exactly at node $a$. 
Instead, it will start at some node $a^{'} = (\bar{x}_a, \hat{P}_{a\m}^{'}, \tilde{P}_{a\m}^{'})$ that satisfies $\hat{P}_{a\m}^{'} \preceq \hat{P}_{a\m}$ and $\tilde{P}_{a\m}^{'} \preceq \tilde{P}_{a\m}$.
Next, we show that by applying the pre-computed feedback gain $K$, the terminal covariance still satisfies $\hat{P}_{N}^{'} \preceq \hat{P}_{b}$ for the new covariances $\hat{P}_{a\m}^{'}$ and $\tilde{P}_{a\m}^{'}$.
Before verifying this result, we provide a property of the covariance of the Kalman filter.
Given an initial estimation error covariance $\tilde{P}_{0\m}$, we obtain a sequence of error covariances $( \tilde{P}_{k\m})_{k=1}^{N}$ using the Kalman filter updates (\ref{KFupdate}). 
Similarly, given the new initial error covariance $\tilde{P}_{0\m}^{'}$ which satisfies $\tilde{P}_{0\m}^{'}\preceq\tilde{P}_{0\m}$, we obtain a new sequence of error covariances $( \tilde{P}_{k\m}^{'})_{k=0}^{N}$.
We have the following lemma.

\begin{lemma} \label{Lemma1}
If $\tilde{P}_{0\m}^{'}\preceq\tilde{P}_{0\m}$, then $\tilde{P}_{k\m}^{'}\preceq\tilde{P}_{k\m}$, for all $k = 0, \cdots, N$.
\end{lemma}
The proof of this lemma is omitted. 
A proof of a similar result can be found in \cite{Bry2011rapidly}.
From Lemma~\ref{Lemma1}, it is straightforward to show that, if $\tilde{P}_{0\m}^{'}\preceq\tilde{P}_{0\m}$, we also have $\tilde{P}_{k}^{'}\preceq\tilde{P}_k$ for all $k = 0, \ldots, N$.

\begin{proposition} \label{Prop4}
Consider a path on the CS-BRM roadmap, where the initial node of the path is denoted by $(\bar{x}_i, \hat{P}_i, \tilde{P}_i)$ and the final node of the path is denoted by $(\bar{x}_j, \hat{P}_j, \tilde{P}_j)$.
Starting from the initial node and following this path by applying the sequence of edge controllers, the robot will arrive at a belief node $(\bar{x}, \hat{P}, \tilde{P})$ such that $\bar{x} = \bar{x}_j$, $\hat{P} \preceq \hat{P}_{j}$, and $\tilde{P} \preceq \tilde{P}_{j}$.
% \begin{proof}
% See Appendix \ref{AppendixForProp4}.
% \end{proof}
\end{proposition}

The proof of this proposition is straightforward and is omitted.
Proposition~\ref{Prop4} guarantees that, when planning on the CS-BRM roadmap, the covariance at each arrived node is always smaller than the assigned fixed covariance at the corresponding node of the CS-BRM roadmap.

\section{Numerical Examples} \label{Sec:Implementation}

In this section, we first illustrate our theoretical results for the motion planning problem of a 2-D double integrator.  
We compare our method with the SLQG-FIRM algorithm \cite{Agha-Mohammadi2014FIRM}, and we show that our method overcomes some of the limitations of SLQG-FIRM.
Subsequently, the problem of a fixed-wing aerial vehicle in a 3-D environment is studied.

\subsection{2-D Double Integrator} \label{sec:2Dexample}

A 2-D double integrator is a linear system with a 4-dimensional state space given by its position and velocity, and a 2-dimensional control space given by its acceleration. 

\begin{figure}[ht]
    \centering
    \begin{subfigure}[b]{0.42\columnwidth}
         \centering
         \includegraphics[width=1\columnwidth]{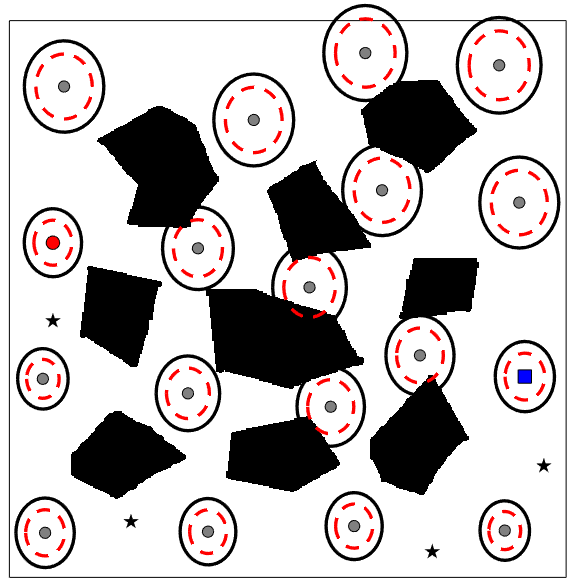}
         \caption{}
     \end{subfigure}
     \begin{subfigure}[b]{0.42\columnwidth}
         \centering
         \includegraphics[width=1\columnwidth]{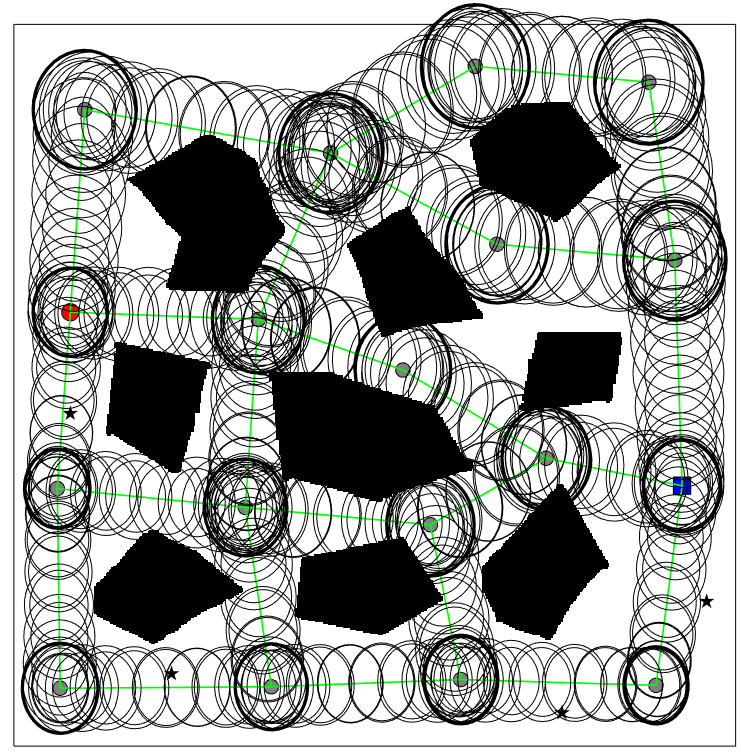}
         \caption{}
     \end{subfigure}
    \caption{(a) Sampled CS-BRM nodes. 
    %The gray dots are the state means $\bar{x}$ of the nodes. The red dot is the state mean of the starting node and the blue square is the state mean of the goal node. 
    %The black ellipses around the state represent the $3\sigma$ confidence intervals of the position covariances, and the red dash ellipses correspond to the prior estimation error covariances. 
    %The goal of the planning problem is to find the optimal path along with the control policy to go from the starting node to the goal node.
    (b) CS-BRM of the 2-D double integrator. The green lines are the mean trajectories between the nodes. The gray ellipses are the $3\sigma$ confidence intervals of the covariances of the positions. 
    %The roadmap is a directed graph.
    }
    \label{2Dmap1}
\end{figure}

The environment considered is shown in Figure~\ref{2Dmap1}.
There are $\ell$ landmarks placed in the environment, which are represented by black stars.
The black polygonal shapes represent the obstacles.
The agent observes all landmarks and obtains estimates of its position at all time steps.
The agent achieves better position estimates when it is closer to the landmarks. 
Let %the location of the $j^{th}$ landmark be given by $L_j$ and 
the Euclidean distance between the position of the 2-D double integrator and the $j^{th}$ landmarks be given by $d_j$.
Then, the $j^{th}$ position measurement corresponding to landmark $j$ is
\begin{equation}
\begin{split}
    {}^j\!y = [x^{(1)} \ x^{(2)}]^\top + \eta_p d_j v_p, \quad j=1,2,\cdots,\ell,
\end{split}
\end{equation}
where $\eta_p $ is a parameter related to the intensity of the position measurement noise that is set to $0.1$, and $v_p$ is a two-dimensional standard Gaussian random vector.
The velocity measurement is given by 
\begin{equation}
\begin{split}
    y_v = [x^{(3)} \ x^{(4)}]^\top + \eta_v v_v,
\end{split}
\end{equation}
where $\eta_v $ is a parameter related to the intensity of the velocity measurement noise that is set to $0.2$, and $v_v$ is a two-dimensional standard Gaussian random vector.
%In this case, the velocity is measured using onboard sensors and does not depend on the landmarks. 
Thus, the total measurement vector $y$ is a $2 \ell + 2$ dimensional vector, composed of $\ell$ position measurements and one velocity measurement.

The sampled CS-BRM nodes are shown in Figure~\ref{2Dmap1}(a). 
The velocities are restricted to be zero for the purpose of comparison with SLQG-FIRM.
The gray dots are the state means of the nodes. The red dot shows the position of the starting node and the blue square shows the position of the goal node.
The black ellipses around the state represent the $3\sigma$ confidence intervals of the position covariances. The red dash ellipses correspond to the prior estimation error covariances. 
Following Algorithm \ref{alg:CSBRM}, the edge controllers are computed using the covariance steering controller for each edge and the collision cost is computed using Monte Carlo simulations.
The final CS-BRM is given in Figure~\ref{2Dmap1}(b).

%Both $\hat{P}_{\m}$ and $\tilde{P}_{\m}$ are sampled from the positive definite matrix space.

%In the covariance steering method introduced in Section \ref{Sec:CovSteering}, the number of time-steps $N$ for a particular edge, and the time step size $\Delta t$ are design variables chosen by the user. In this example, we choose $\Delta t = 0.2$ s. 
%The time duration of the edge is chosen based on the Euclidean distance between the state mean of the two endpoints of that edge. By specifying a desired average speed of the robot, we approximate the time duration of the edge and then obtain the number of time-steps $N$ of this edge.
%The desired average speed is a design parameter and is set to be 4~m/s.
% One of the more sophisticated ways to choose these design variables is to search the design space. For example, search different time duration and select the one with the smallest mean control cost. 
%The final reference trajectory obtained using Algorithm~\ref{alg:CNT} is also the mean trajectory of the corresponding edge.

%The edge cost in CS-BRM is the combination of the mean control cost, the covariance control cost, and the cost from the probability of collision.
After the CS-BRM is built, the path planning problem on CS-BRM is the same as the problem of planning using a PRM, which can be easily solved using a graph search algorithm. 
The difference between the CS-BRM and PRM is that the edge cost in CS-BRM is specifically designed to deal with dynamical system models and uncertainties, and the transition between two nodes of CS-BRM is achieved using covariance steering as the edge controller. 

\begin{figure}[ht]
    \centering
     \begin{subfigure}[b]{0.42\columnwidth}
         \centering
         \includegraphics[width=1\columnwidth]{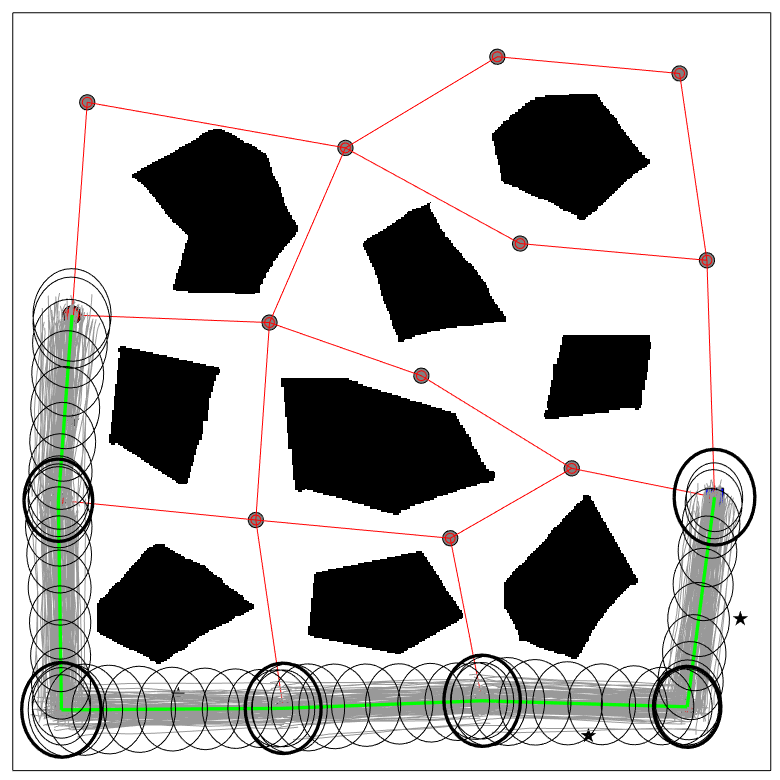}
         \caption{}
     \end{subfigure}
          \begin{subfigure}[b]{0.42\columnwidth}
         \centering
         \includegraphics[width=1\columnwidth]{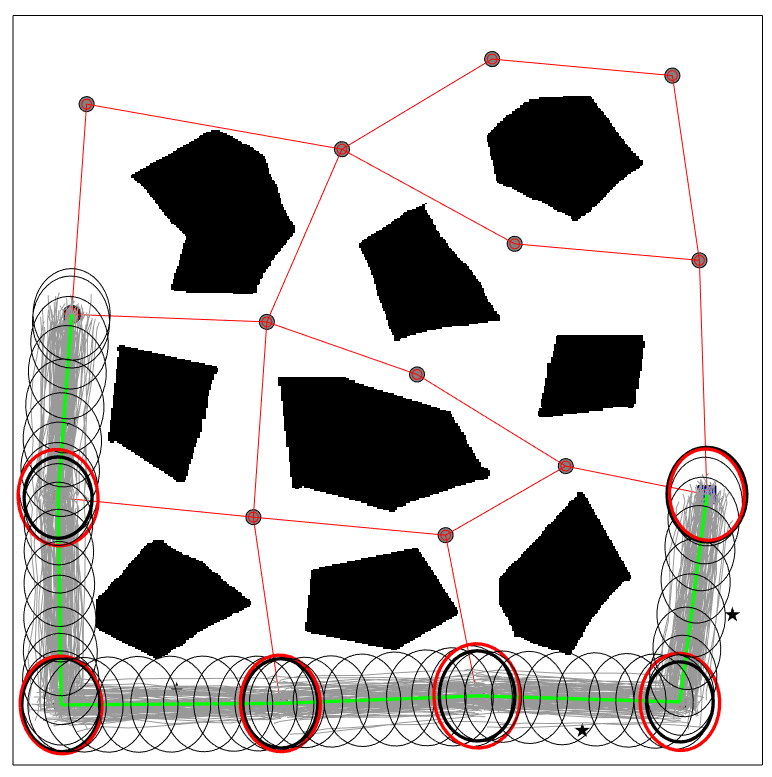}
         \caption{}
     \end{subfigure}
    \caption{Results of sampling stationary belief nodes. (a) Path planned using CS-BRM; (b) Path planned using SLQG-FIRM. 
    %The red ellipses are the state covariances at the last step of the time-varying LQG controller. The time-varying LQG controller is switched to the SLQG controller until the state covariance converges to a value that is smaller than the state covariance of the belief nodes (bold black ellipses).
    }
    \label{2DmapPlan}
\end{figure}

The planning results of CS-BRM and SLQG-FIRM are given in Figure~\ref{2DmapPlan}. 
The paths goes through the same set of belief nodes for this example.
In Figure~\ref{2DmapPlan}(b), the red ellipses are the state covariances at the last step of the time-varying LQG controller. 
As we see, each one of the ellipses is larger than the corresponding state covariance of the intermediate belief node (bold black ellipses). 
Thus, a converging step is required at every intermediate node.
The time-varying LQG controller is switched to the SLQG controller until the state covariance converges to a value that is smaller than the state covariance of the belief nodes (bold black ellipses).

\begin{figure}[ht]
    \centering
    \begin{subfigure}[b]{0.42\columnwidth}
         \centering
         \includegraphics[width=1\columnwidth]{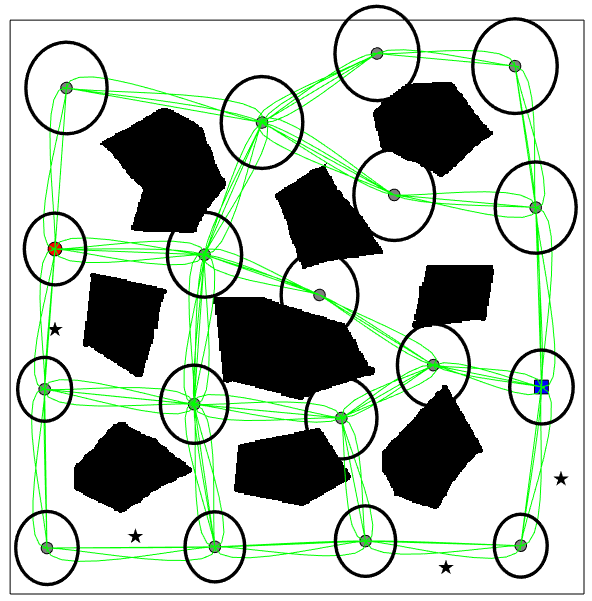}
         \caption{}
     \end{subfigure}
     \begin{subfigure}[b]{0.42\columnwidth}
         \centering
         \includegraphics[width=1\columnwidth]{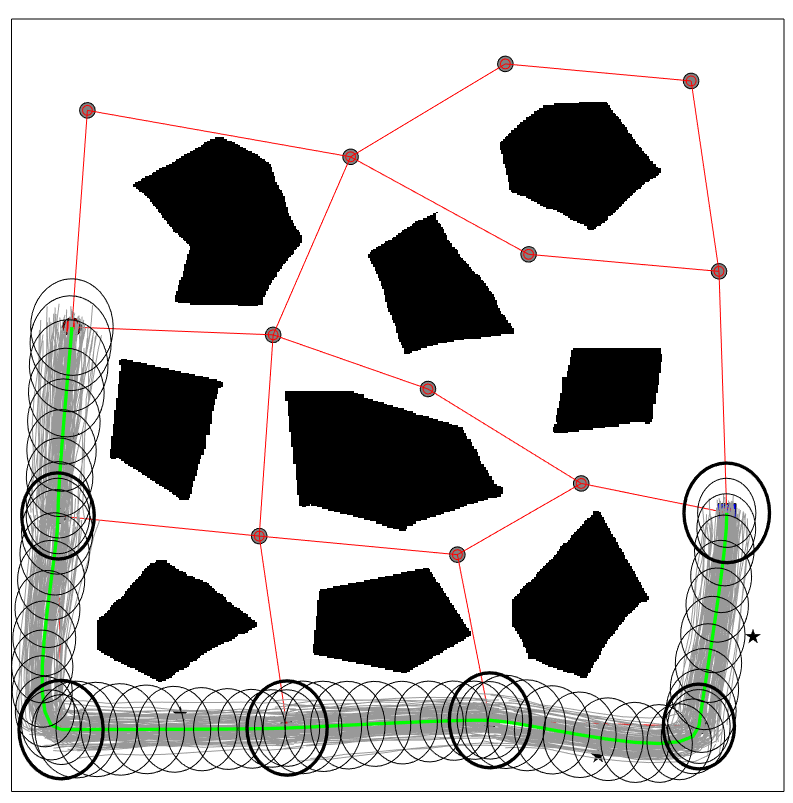}
         \caption{}
     \end{subfigure}
    \caption{(a) CS-BRM results of sampling nonstationary nodes.
    The green lines are the mean trajectories of the edges.
    (b) Planning results. By exploring the velocity space, the robot does not need to stop at every node. The path cost is much smaller compared to Figures \ref{2DmapPlan}(a) and \ref{2DmapPlan}(b).}
    \label{differentVelBRM}
\end{figure}

The CS-BRM that samples nonstationary nodes is shown in Figure~\ref{differentVelBRM}(a). 
For clarity, only the mean trajectories of the edges are shown (green lines). 
The environment, the sampled mean positions, and the sampled covariances are all the same as those in Figure \ref{2DmapPlan}(a) and \ref{2DmapPlan}(b). 
The only difference is that multiple velocities are sampled at each position in Figure \ref{differentVelBRM}(a).  
The planned path is shown in Figure \ref{differentVelBRM}(b). 
The cost of the planned path is $104.87$, which is much lower than the cost of the paths in Figure~\ref{2DmapPlan}(a) and Figure~\ref{2DmapPlan}(b), which are $239.36$ and $240.47$ respectively.
The decrease in the path cost is due to the decrease of the mean control cost.
In Figure~\ref{2DmapPlan}(b), the robot has to stop (zero velocity) at every intermediate node.  
With the proposed method, the robot goes through each intermediate node smoothly, which results in a more efficient path. 

\subsection{Fixed-Wing Aircraft}

In this section, we apply the proposed method on a nonlinear example, namely, planning for a fixed-wing aerial vehicle~\cite{Owen2015Implementing}.
The discrete-time system model is given by
\begin{equation}
\begin{split}
x_{k+1} &= x_k + V_k \cos\psi_k \cos\gamma_k \Delta t + g_{1k} w_{1k}, \\
y_{k+1} &= y_k + V_k \sin\psi_k \cos\gamma_k \Delta t + g_{2k} w_{2k}, \\
z_{k+1} &= z_k + V_k \sin\gamma_k \Delta t + g_{3k} w_{3k}, \\
\psi_{k+1} &= \psi_k + \frac{g}{V_k} \tan\phi_k \Delta t + g_{4k} w_{4k},
\end{split}
\end{equation}
where the 4-dimensional state space is given by $[x \ y \ z \ \psi]^\top$, where $[x \ y \ z]^\top$ is the 3-D position of the vehicle, and $\psi$ is the heading angle. 
The 3-D control input space is given by $[V \ \gamma \ \phi]^\top$, where $V$ is the air speed, $\gamma$ is the flight-path angle, and $\phi$ is the bank angle, $\Delta t$ is the time step size,
$w_{ik}$, $i=1,2,3,4$, are standard Gaussian random variables, $g_{1k}$, $g_{2k}$, $g_{3k}$, and $g_{4k}$ are multipliers correspond to the magnitude of the noise. 
Their values are all set to be $0.1$.
%The system is linearized along a nominal trajectory to obtain a discrete, linear time-varying model.

Similar to the 2-D double integrator example, several landmarks are placed in the environment.
%We assume that the vehicle can observe all landmarks and obtain estimates of the state at all time steps. The vehicle achieves better state estimates when it is closer to the landmarks.
Let %the location of the $j^{th}$ landmark be given by $L_j$ and 
the Euclidean distance between the 3D position of the vehicle and the $j^{th}$ landmarks be given by $d_j$.
Then, the $j^{th}$ position measurement corresponding to landmark $j$ is
\begin{equation}
    {}^j\!y = [x \ y \ z \ \phi]^\top + \eta d_j v, \quad j=1,2,\ldots,\ell,
\end{equation}
where $\eta $ is a parameter related to the intensity of the noise of the measurement
and is set to $0.05$, and $v$ is a 4-dimensional standard Gaussian random vector.
Thus, the total measurement vector $y$ is a $4 \ell$-dimensional vector, where $\ell$ is the number of landmarks.

\begin{figure}[ht]
    \centering
    \begin{subfigure}[b]{1\columnwidth}
         \centering
         \includegraphics[width=0.49\columnwidth]{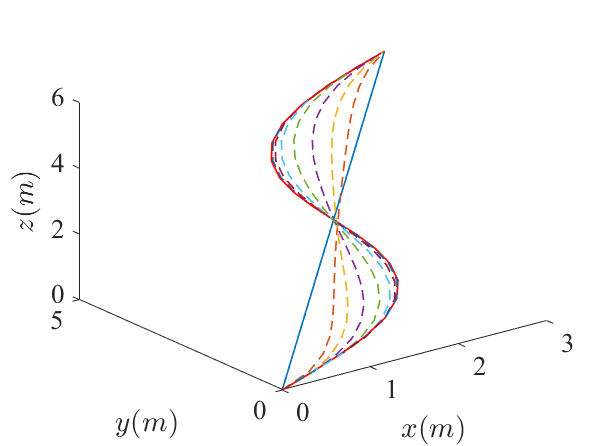}
         \includegraphics[width=0.49\columnwidth]{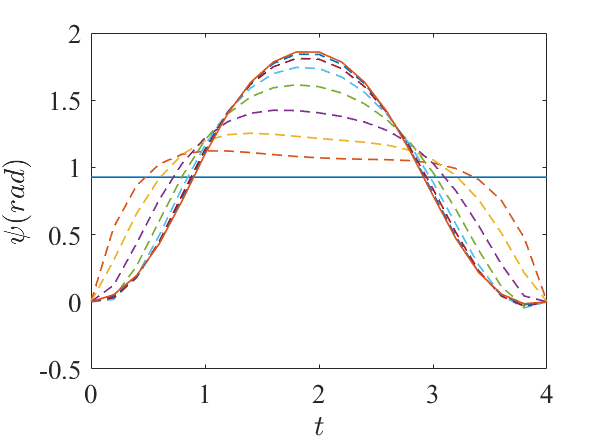}
     \end{subfigure}
    \caption{Generation of a compatible nominal trajectory for the fixed-wing vehicle.
 	 The nominal trajectory and the reference trajectory are initialized as straight lines. 
 	 %The state trajectories are shown in (a) and     	 the control trajectories are shown in (b).     	 The blue solid lines correspond to the initial nominal trajectory (which is also the reference trajectory). The red lines correspond to the final mean trajectory, which is also the compatible nominal trajectory and the reference trajectory.
 	 }
 \label{3DmeanTrajIte}
\end{figure}

Algorithm~\ref{alg:CNT} was used to find a compatible nominal trajectory for the fixed-wing vehicle. 
One example is given in Figure~\ref{3DmeanTrajIte}.
The nominal trajectory and the reference trajectory are initialized as straight lines in all four state dimensions and the three control input dimensions.
The algorithm converged in eight iterations.

\begin{figure}[ht]
    \centering
        \includegraphics[width=0.75\columnwidth]{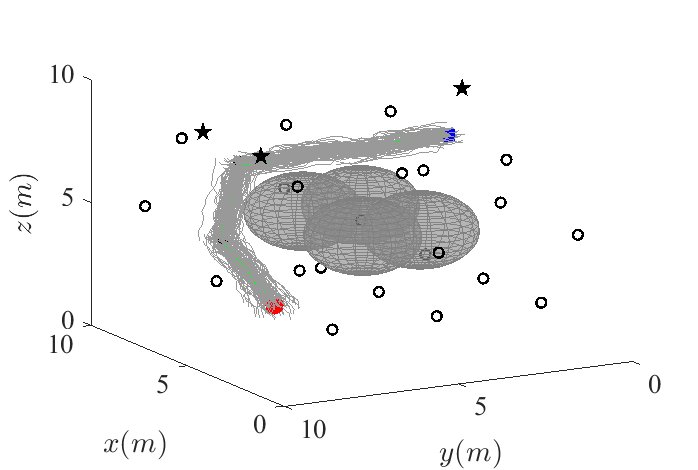}
    \caption{Planning results using the CS-BRM for the fixed-wing vehicle. The gray lines are the trajectories from the Monte-Carlo simulations.
    The algorithm finds the optimal path that trades off between control effort and collision cost. }
    \label{3DCSBRM}
\end{figure}

The planning environment is shown in Figure~\ref{3DCSBRM}. 
The four spheres in the middle of the environment are the obstacles.
The three black stars represent the landmarks.
%The CS-BRM nodes are also shown in Figure~\ref{3DCSBRM}.
The black circles represent the state mean (position) of the nodes.
The red dot shows the position of the starting node and the blue square shows the position of the goal node.
The state means are deterministically chosen to discretize the environment.
The covariances of the nodes are not showed.

The planned path using this CS-BRM is also shown in Figure~\ref{3DCSBRM}. 
Similarly to the example in Section \ref{sec:2Dexample},
instead of flying through the narrow passage between the four obstacles, which resulting a high probability of collision, the algorithm choose a path that trades off between control cost and collision cost while minimizing the total edge costs.

\section{Conclusion}  \label{Sec:Conclusion}

A belief space roadmap (BRM) algorithm is developed in this paper.
The nodes in BRM represent distributions of the state of the system and are sampled in the belief space.
The main idea is to use covariance steering to design the edge controllers of the BRM graph to steer the system from one distribution to another.
Compared to \cite{Agha-Mohammadi2014FIRM}, the proposed method allows sampling nonstationary belief nodes, which has the advantage of more complete exploration of the belief space.
For covariance steering of nonlinear systems, we introduce the concept of compatible nominal trajectories, which aim to better approximate the nonlinear dynamics through successive linearization.
We also propose an efficient algorithm to compute compatible nominal trajectories.
Compared to the standard PRM, the additional computation load comes from the computation of the edge controllers and edge cost evaluations, which, however, are done offline.
By explicitly incorporating motion and observation uncertainties, we show that the proposed CS-BRM algorithm generates efficient motion plans that take into account both the control effort and collision probability.

\section*{Acknowledgments}
This work has been supported by NSF awards IIS-1617630 and IIS-2008695, NASA NSTRF Fellowship 80NSSC17K0093, and ARL under CRA DCIST W911NF-17-2-0181.
The authors would also like to thank Dipankar Maity for many insightful discussions regarding the output-feedback covariance steering problem. 

\bibliographystyle{ieeetran}
\bibliography{refs}

\end{document}